\documentclass[10pt,twocolumn,letterpaper]{article}

\usepackage{iccv}
\usepackage{times}
\usepackage{epsfig}
\usepackage{graphicx}
\usepackage{amsmath}
\usepackage{amssymb}
\usepackage{booktabs}
\usepackage[table]{xcolor}
\usepackage{pifont}
\usepackage{multirow}
\usepackage{color}
\definecolor{mybar}{rgb}{1.0, 0.6}

\usepackage[pagebackref=true,breaklinks=true,letterpaper=true,colorlinks,bookmarks=false]{hyperref}

\iccvfinalcopy 

\begin{document}

\title{GLH-Water: A Large-Scale Dataset for Global Surface Water Detection in Large-Size Very-High-Resolution Satellite Imagery}

\author{Yansheng Li \thanks{corresponding author.} \qquad Bo Dang \qquad Wanchun Li \qquad Yongjun Zhang\\
Wuhan University, Wuhan, China\\
{\tt\small \{yansheng.li, bodang, wanchun.li, zhangyj\}@whu.edu.cn}}

\maketitle

\begin{abstract}
   Global surface water detection in very-high-resolution (VHR) satellite imagery can directly serve major applications such as refined flood mapping and water resource assessment. Although achievements have been made in detecting surface water in small-size satellite images corresponding to local geographic scales, datasets and methods suitable for mapping and analyzing global surface water have yet to be explored. To encourage the development of this task and facilitate the implementation of relevant applications, we propose the GLH-water dataset that consists of 250 satellite images and manually labeled surface water annotations that are distributed globally and contain water bodies exhibiting a wide variety of types (\eg~, rivers, lakes, and ponds in forests, irrigated fields, bare areas, and urban areas). Each image is of the size 12,800 $\times$ 12,800 pixels at 0.3 meter spatial resolution. To build a benchmark for GLH-water, we perform extensive experiments employing representative surface water detection models, popular semantic segmentation models, and ultra-high resolution segmentation models. Furthermore, we also design a strong baseline with the novel pyramid consistency loss (PCL) to initially explore this challenge. Finally, we implement the cross-dataset and pilot area generalization experiments, and the superior performance illustrates the strong generalization and practical application of \textit{GLH-water}. The dataset is available at \url{https://jack-bo1220.github.io/project/GLH-water.html}.
\end{abstract}

\begin{figure}[t]
  \centering
   \includegraphics[width=\linewidth]{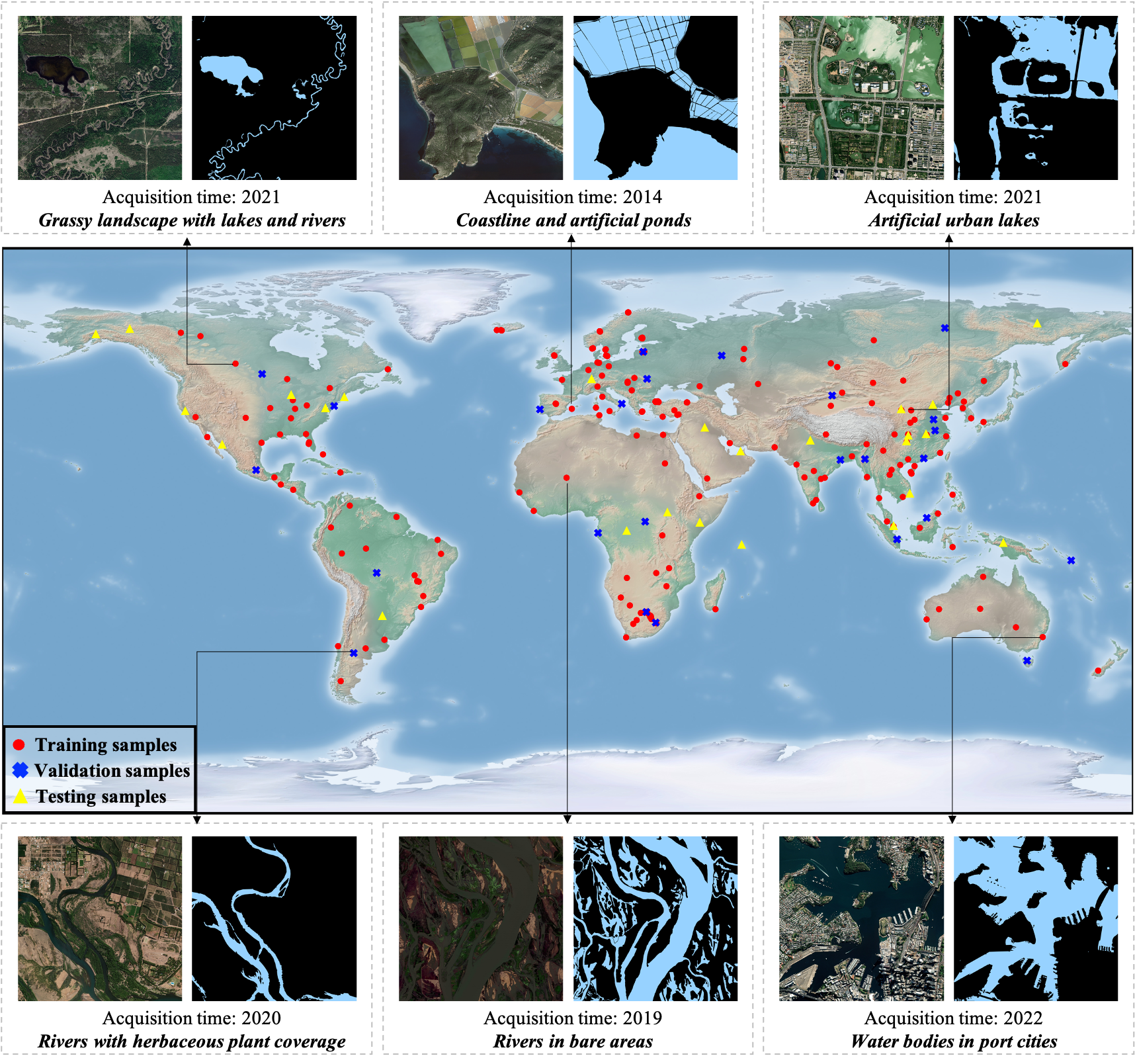}
   \caption{\textbf{Visualization of the \textit{GLH-water} dataset.} We show the geographical coverage of the samples. Several examples from different continents are selected and their image acquisition times and scene descriptions are provided.}
   \label{fig:1}
   \vspace{-1.0em}
\end{figure}

\section{Introduction}

As one of the fundamental components of the Earth's natural ecosystem, surface water plays a critical role in maintaining biodiversity, ecological balance, and the development of human societies \cite{vorosmarty2010global}. Due to its wide spatial and temporal distribution, using satellite imagery to detect and map the global surface water is a feasible and convenient method, leading to promising breakthroughs that are applied in the flood mapping \cite{mateo2021towards, wieland2023semantic}, surface water changes \cite{donchyts2016earth, pekel2016high}, and other assessments of water resources \cite{wang2020gainers, grill2019mapping}.

Compared to synthetic aperture radar (SAR) imagery, very-high-resolution (VHR) optical satellite imagery (Ground Sampling Distance, GSD\textless5m) has the advantage of providing clearer texture and detail information about water. In contrast, medium- and low-resolution optical images can only detect large surface water bodies, while small-scale water bodies and their intricate details can only be captured by VHR optical satellite imagery.

\begin{table*}[]
\centering
\renewcommand\arraystretch{1.2}
\resizebox{175mm}{13mm}{
\begin{tabular}{lccccccccccc}
\toprule
Dataset                                                                                & DeepWaterMap \cite{isikdogan2019seeing} & ESKWB \cite{luo2021applicable} & SWED \cite{seale2022coastline} & \begin{tabular}[c]{@{}c@{}}2020 GF\\ challenge \cite{sun2021automated}\end{tabular}  & Landcover.ai \cite{Boguszewski_2021_CVPR} & Agriculture-vision \cite{Chiu_2020_CVPR} & LoveDA \cite{wang2021loveda} & FBP \cite{tong2023enabling} & DynamicEarthNet \cite{Toker_2022_CVPR} & OpenEarthMap \cite{xia_2023_openearthmap} & \textbf{GLH-water (ours)} \\ \midrule
\begin{tabular}[c]{@{}l@{}}Dedicated to surface \\ water detection tasks\end{tabular} & \ding{52}            & \ding{52}     & \ding{52}    & \ding{52}                                                           &              &                    &        &     &           &      &  \ding{52}                           \\
Global sample                                                                          & \ding{52}            & \ding{52}    & \ding{52}    &                                                             &              &                    &        &     & \ding{52}             & \ding{52} &  \ding{52}                           \\
VHR                                                                &              &       &      & \ding{52}                                                           & \ding{52}            & \ding{52}                  & \ding{52}      & \ding{52}   & \ding{52}         & \ding{52}      &  \ding{52}                           \\
Large-size imagery                                                                     &              &       &      &                                                             & \ding{52}            &                    &        & \ding{52}   &            &     & \ding{52}                           \\
Large-scale dataset                                                                    & \ding{52}            &       &      &                                                             &              &                    &        &     & \ding{52}         &      & \ding{52}                           \\ \bottomrule
\end{tabular}}
\vspace{0.0em}
\caption{\textbf{Characteristics of \textit{GLH-water} dataset compared with other public datasets.} The analysis demonstrates that the \textit{GLH-water} is the first large-scale dataset for global surface water detection in large-size VHR optical satellite imagery. \texttt{VHR} means GSD\textless5m. \texttt{Large-size imagery} means that size of each image is larger than 10,000 $\times$ 10,000 pixels. \texttt{Large-scale dataset} means the total number of images is equivalent to more than 100,000 tiles of 512 $\times$ 512 pixels in size.}
\label{tab:table1}
\vspace{-1.0em}
\end{table*}

To our knowledge, no publicly suitable data has been proposed to effectively facilitate the training and evaluation of surface water detection methods in global VHR optical imagery, as highlighted in Table \ref{tab:table1}, which seriously hinders the advancement of global VHR surface water body mapping tasks. In addition, the size of VHR satellite images of the whole scene is typically large (\eg~, the standard scene size of GaoFen-2 satellite images is approximately 6,000 $\times$ 6,000 pixels). Extracting complete and continuous water bodies from large-size images poses not only a greater challenge that remains unexplored, but also is closer to practical applications such as large-scale surface water mapping.

To promote research on this challenging task and to support applications like higher-resolution global surface water mapping, we propose \textit{GLH-water}, a large-scale dataset for global surface water detection in large-size VHR optical satellite imagery. We collect 250 VHR (GSD=0.3m) optical satellite images of 12,800 $\times$ 12,800 pixels containing various water types from the whole world, and create accurate annotations through manual labeling and expert inspection, as shown in Figure~\ref{fig:1}. \textit{GLH-water} significantly differs from other existing datasets analyzed in Table \ref{tab:table1}, with advantages focusing on the following aspects: (1) large-size images; (2) a large number of samples; (3) extensive geographical coverage of the samples; (4) broad temporal span of image acquisition; (5) inclusion of diverse types of surface water. These variations and traits give \textit{GLH-water} its uniqueness and present new challenges for detecting surface water from large-size VHR optical satellite imagery.

Furthermore, to evaluate the dataset and explore this challenging task, we evaluate the performance of representative surface water detection models, well-performing general semantic segmentation models and ultra-high resolution segmentation methods on \textit{GLH-water}, and combine their metrics to create a benchmark. Motivated by multi-layer visual field difference of the image pyramid and the topological continuity of surface water in large-size satellite images, we propose a strong baseline with the new pyramid consistency loss (PCL) to offer a promising pipeline for this challenge. Finally, we conduct extensive experiments to demonstrate the strong generalization of \textit{GLH-water} and its potential application value. In summary, our contributions are as follows:
\begin{itemize}
  \vspace{-0.5em}
  \item We present the first large-scale dataset for global surface water detection in large-size VHR optical satellite imagery, which offers significant advantages for tackling the challenge of detecting surface water from large-size satellite images that remains unexplored. 
  \vspace{-0.5em}
  \item We evaluate a variety of semantic segmentation models on \textit{GLH-water}, including representative surface water detection models, advanced segmentation models, and ultra-high resolution segmentation algorithms, which can serve as a benchmark for the development of future methods.
  \vspace{-0.5em}
  \item We further propose a novel strong baseline with the PCL, which yields significant improvements and suggests that it will be a competitive pipeline for future development.
\end{itemize}

In the future, \textit{GLH-water} and strong baseline we proposed are also expected to provide reliable training data and models for manufacturing the global high-resolution surface water map.

\section{Related work}
\subsection{Relevant datasets}
\noindent\textbf{Surface water detection datasets.} The upper portion of Table~\ref{tab:table2} displays existing datasets specifically designed for detecting surface water bodies. The majority of currently available datasets \cite{hu2022multi,isikdogan2019seeing,luo2021applicable,seale2022coastline} for surface water detection based on optical satellite imagery exhibit only low to medium spatial resolution, as seen with Landsat-8 and Sentinel-2 images. The resolution limitations of the images result in the blurring and indistinguishability of small rivers and lakes. To track intricate surface water systems, research endeavors direct their experimental focus towards commercial satellites that offer high resolution of up to 1m or even 0.3m, such as GeoEye and WorldView \cite{moortgat2022deep,wieland2023semantic}. Regrettably, owing to the policy constraints of commercial satellites, these datasets will not be made available to the public community and only be allowed to conduct private experimental validation for their owners. In contrast, our \textit{GLH-water} is the first publicly available large-scale dataset for surface water detection from global VHR optical satellite imagery.

\begin{table*}[]
\centering
\renewcommand\arraystretch{1.2}
\resizebox{175mm}{44mm}
{\begin{tabular}{ccccccccc}
\toprule
Dataset                    & \# Images            & \# Channels & Image size (pixels)                                            & GSD (m)   & \# Labeled pixels (billion)   & Sources                                                             & Geographic coverage & Acquisition     \\ \toprule
\multicolumn{8}{l}{\textit{\textbf{Dedicated surface water detection dataset}}}                                                                                                                                                                               \\
DeepWaterMap \cite{isikdogan2019seeing}               & \textgreater{}140000 & 6           & -                                                              & 30        & -   & Landsat-8                                                           & Globe               & Public          \\
WSD \cite{hu2022multi}                       & 16320                & 3        & 256×256                                                        & 30  & 1.07         & Landsat-8                                                           & -                   &  Private         \\
ESKWB \cite{luo2021applicable}                      & 95                   & 6           & 545$\sim$1432×625$\sim$1527                                    & 10       & 0.11    & Sentinel-2                                                          & Globe               & Public          \\
SWED \cite{seale2022coastline}                      & 1862                 & 12          & 256×256                                                        & 10    & 0.12       & Sentinel-2                                                          & Globe               & Public          \\
SWB                        & 2841                 & 3           & 57$\sim$5292×57$\sim$6767                                      & 10       & -   & Sentinel-2                                                          & -                   & Public          \\
2020 GF challenge \cite{sun2021automated}          & 1000                 & 3           & 492$\sim$2000×492$\sim$2000                                    & 1 to 4   & 0.24    & GaoFen-2                                                            & -                   & Public          \\
Moortgat \etal \cite{moortgat2022deep}            & 142                  & 4/8         & 10000×10000                                                    & 1.2     & 14.2     & GeoEye etc.                                                         & Arctic              & Private         \\
Wieland \etal \cite{wieland2023semantic}             & 1120                 & 4           & 2048×2048                                                      & 0.8    & 4.69      & IKONOS etc.                                                         & Globe               & Private         \\ \midrule
\multicolumn{8}{l}{\textit{\textbf{LULC dataset (including water bodies)}}}                                                                                                                                                                                   \\
DeepGlobe \cite{Demir_2018_CVPR_Workshops}                  & 803                  & 3           & 2448×2448                                                      & 0.5     & 4.81     & DigitalGlobe                                                        & -                   & Public          \\
GID \cite{tong2020land}                        & 10/150               & 3/4         & 7200×6800                                                      & 4   & 7.34         & GaoFen-2                                                            & China               & Public          \\
Landcover.ai \cite{Boguszewski_2021_CVPR}              & 41                   & 3           & \begin{tabular}[c]{@{}c@{}}9000×9500/\\ 4200×4700\end{tabular} & 0.25/0.5  & 3.50   & \begin{tabular}[c]{@{}c@{}}Public Geodetic \\ Resource\end{tabular} & Poland              & Public          \\
Agriculture-Vision \cite{Chiu_2020_CVPR}        & 94986                & 4           & 512×512                                                        & 0.1/0.15/0.2 & 24.9  & UAV camera                                                          & United States       & Public          \\
LoveDA \cite{wang2021loveda}                    & 5987                 & 3           & 1024×1024                                                      & 0.3        & 6.27  & Google Earth                                                        & China               & Public          \\
FBP \cite{tong2023enabling}                       & 150                  & 4           & 7200×6800                                                      & 4     & 7.34       & GaoFen-2                                                            & China               & Public          \\
UrbanWatch \cite{zhang2022urbanwatch}                & 200                  & 4           & 512×512                                                        & 1     & 0.05       & NAIP                                                                & United States       & Public          \\
DynamicEarthNet \cite{Toker_2022_CVPR}           & 54750                & 4           & 1024×1024                                                      & 3     & 57.40       & PlanetFusion                                                        & Globe               & Public          \\
OpenEarthMap \cite{xia_2023_openearthmap}                     & 5000                     & 3         & 1024×1024                                                               & 0.25-0.5    & 5.24    & -                                                     & Global 97 regions               & Public          \\ \midrule
\textbf{GLH-water (our)} & \textbf{250}         & \textbf{3}  & \textbf{12800×12800}                                           & \textbf{0.3} & \textbf{40.96} & \textbf{Google Earth}                                               & \textbf{Globe}      & \textbf{Public} \\ \bottomrule
\end{tabular}}
\vspace{0.0em}
\caption{\textbf{Comparison among \textit{GLH-water} and other relevant datasets.} All datasets are compared on number of images and channels, image size, spatial resolution (GSD), data sources, geographic coverage and acquisition.}
\label{tab:table2}
\vspace{-1.0em}
\end{table*}

\noindent\textbf{Land use and land cover (LULC) datasets.} As a fundamental application in the field of remote sensing, datasets related to LULC are extensively created and utilized, and they commonly encompass the class of water. However, they inadequately fulfill the need of tasks such as refined global surface water detection and mapping. The reason is that none of them can simultaneously satisfy the trinity of global sampling, VHR, and large-size imagery, as demonstrated in the lower portion of Table \ref{tab:table2}. For example, FBP \cite{tong2023enabling} originates from the VHR and large-size images of GaoFen-2 satellite. However, it is confined to a limited selection of cities in China, specifically Beijing, and migration of the trained model to other regions presents a challenge. Similarly, the geographic coverage of the Landcover.ai \cite{Boguszewski_2021_CVPR} is restricted to Poland, and the trained models are limited in scale and generalizability. The image size of DynamicEarthNet \cite{Toker_2022_CVPR} is only 1,024×1,024 pixels, insufficient to effectively portray the distribution of water bodies, a characteristic with notable geospatial continuity. In contrast, our \textit{GLH-water} offers distinct advantages such as global sampling, VHR, large-size imagery. These attributes are essential for executing surface water mapping tasks on a global scale.

\subsection{Relevant methods}
\label{sec2.2}
\noindent\textbf{Surface water detection methods based on non-deep learning algorithms.} Normalized Difference Water Index (NDWI) \cite{mcfeeters1996use}, Modified Normalized Difference Water Index (MNDWI) \cite{xu2006modification}, High Resolution Water Index (HRWI) \cite{yao2015high}, Two-step Urban Water Index (TSUWI) \cite{wu2018two}, and other threshold-based water indices are commonly proposed and implemented in initial studies. However, their reliance on spectral information result in a lack of consideration for the spatial information present within the images. Additionally, shallow classifiers such as Support Vector Machine (SVM) and modified Statistical Region Merging (SRM) are employed and show significant improvements \cite{zhang2018automatic,wu2018two,sui2013automatic}.

\noindent\textbf{Surface water detection methods based on deep learning algorithms.} The VHR optical satellite imagery comprises of a range of water bodies, including but not limited to, rivers, lakes, and ponds with diverse sizes and shapes \cite{li2022water}. Numerous studies \cite{duan2019multiscale,yu2021self,cui2020sanet,kang2021multi,yuan2021deep} aim to enhance the identification of intricate water bodies by optimizing the deep learning network, thereby enabling more efficient utilization of the multiscale characteristic. The meandering of water body boundaries constitutes a critical hindrance to the precise segmentation of surface water bodies. Miao \etal \cite{miao2018automatic} devise a loss function to derive accurate water body boundaries, taking into account the distribution of boundary weights. Efficient post-processing algorithms \cite{sun2021automated,chu2019sea} are also effective approaches.

\subsection{Relevant applications}
In the broader context of global water resource monitoring and mapping, scholars create a worldwide 250m land surface water body mask raster data employing MODIS satellite images, grounded on the already existing global land surface water body vector data \cite{carroll2009new}. The European Commission Joint Research Centre (ECJRC) use Landsat-5/7/8 satellite imagery to map global 30m water body data products for the period 1984-2020 \cite{pekel2016high}. Global land cover products which incorporate the water body category are gradually emerging, albeit at a resolution of merely 10m \cite{jun2014open, chen2019stable,karra2021global}. In addition, continuous production and application of regional mapping products with low to medium resolution imagery for water bodies \cite{wang2022improved,li2022systematic,feng2016global} and with high resolution imagery for LULC \cite{Robinson_2019_CVPR,li2022breaking} is ongoing. The inevitable trend in cartography is to continuously enhance the spatial resolution of products, as this demonstrates the benefits of more detailed information. However, the production of global VHR water cover maps remains a challenging task, due to the difficulty of acquiring and organizing VHR satellite data, the absence of publicly available large-scale surface water detection datasets with manual annotation, and the lack of related models that are suitable for large-size images. Our \textit{GLH-water} with strong generalization and proposed strong baseline will fill the aforementioned gap.


\section{The \textit{GLH-water} dataset}
To fill the lack of pertinent datasets and enhance the generalizability of segmentation models in detecting global surface water, we first present the \textit{GLH-water} dataset that contains 250 VHR (GSD=0.3m) satellite images with the size of 12,800 × 12,800 pixels. These images are collected from various locations worldwide and manual annotations are included, as illustrated in Figure~\ref{fig:1}. In the remainder of this section, we provide details on the imagery, annotations, and advantages of our dataset.

\subsection{Images collection and preprocessing}
Using the Google Earth platform, we collect a total of 250 satellite images that are evenly distributed across the globe, each with sizes of 12,800 × 12,800 pixels at 0.3m (19 level) spatial resolution, and encompassing approximately 3,686 $km^{2}$ in geographic coverage. To guarantee the diversity of the dataset, we handpick the geographic coordinates and acquisition time of the sample data to ensure an accurate representation of the various attributes of the global water bodies.

\subsection{Annotation method and inspection}
The annotation labeling process includes three distinct stages: fine labeling, fine checking and correction, and random checking conducted by experts. To conduct fine labeling, we initially employ the ArcGIS software to manually outline the boundaries of water bodies in the collected images. In case of ambiguity in certain areas, we refer to Google Maps of the corresponding areas to aid in the labeling process. To ensure the accuracy of the annotations, during the second phase, we concentrate on validating the boundary accuracy, detecting any possible redundancies and omissions, and rectifying each of the issues that surfaced during the inspection. Finally, the annotations accompanying each large-size image are cropped into 512 x 512 pixel tiles in order to accurately reflect their quality. A second visual inspection is conducted by experts on a randomly selected 15\% of the data. 

After scrutiny and revision, no apparent erros are found in the \textit{GLH-water} dataset, and the standard of annotations is high, which sufficiently fulfills the necessities for the forthcoming algorithmic evaluation. Some annotated samples in our dataset are shown in Figure~\ref{fig:3}.

\subsection{Advantage analysis}
To the best of our knowledge, the \textit{GLH-water} dataset is the first publicly available and largest dedicated datset for global-scale surface water detection from large-size VHR satellite imagery that meets the traits of global sampling, VHR, and large image size. A comparison with other publicly existing datasets is shown in Table \ref{tab:table2}.

\begin{figure*}
	\centering
		\includegraphics[scale=.34]{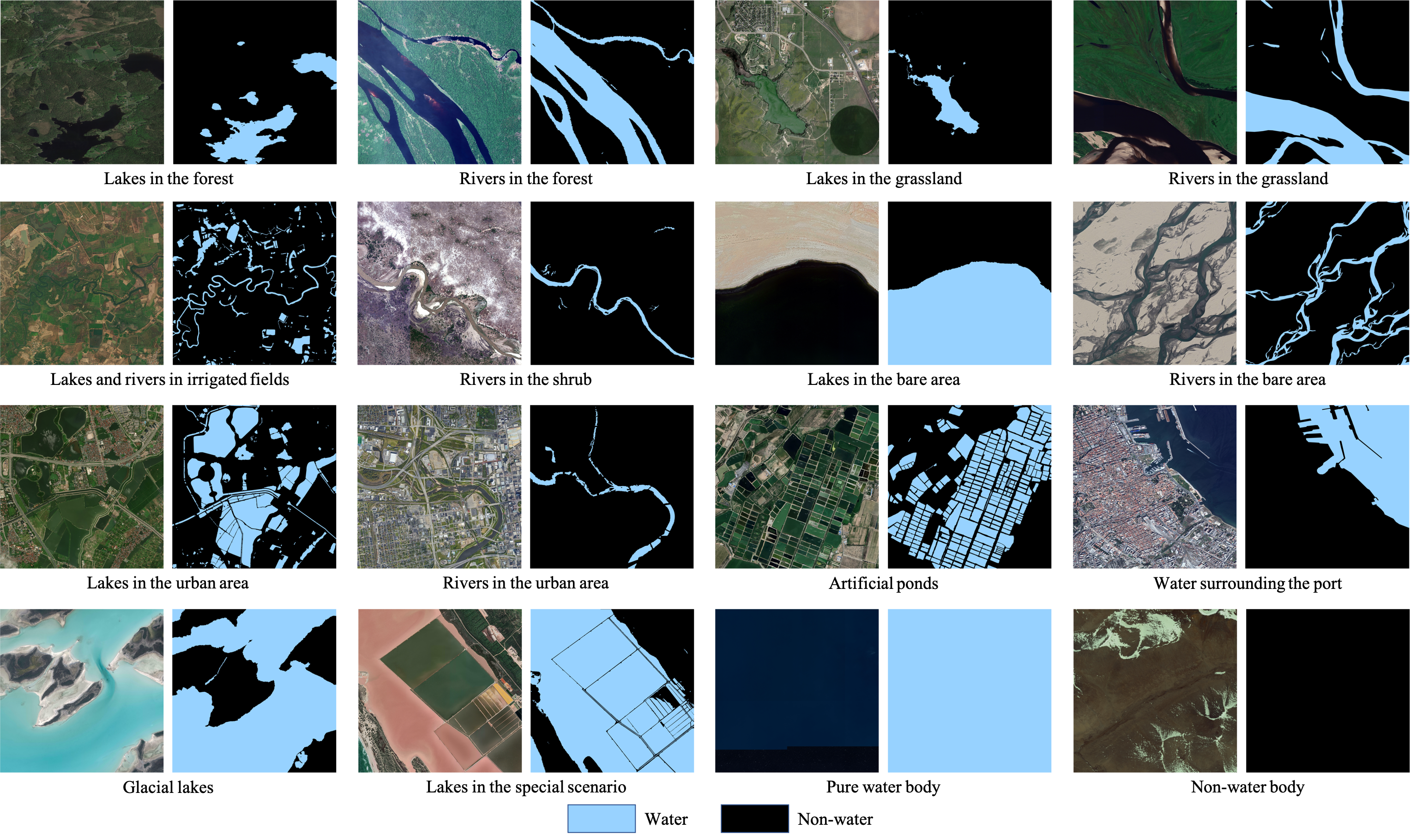}
	    \caption{\textbf{Visualization of various types of surface water bodies in different scenarios on \textit{GLH-water} dataset.} It can comprehensively reflect the diversity of the global surface water system.}
	\label{fig:3}
\vspace{-1.0em}
\end{figure*}

Specifically, our \textit{GLH-water} dataset has five remarkable and important advantages:
\begin{itemize}
  \vspace{-0.5em}
  \item \textbf{The size of samples is large.} The size of each image is up to 12,800 $\times$ 12,800 pixels, which is more in line with the size of a whole scene image acquired by satellite. It is also a challenge for existing methods.
  \vspace{-0.5em}
  \item \textbf{Inclusion of a large number of samples.} After non-overlapping cropping, a total of 156,250 tiles of 512 $\times$ 512 pixels in size and more than 40.96 billion labeled pixels are included, which is the largest dataset for global surface water detection from large-size VHR satellite imagery.
  \vspace{-0.5em}
  \item \textbf{The geographical coverage of the samples is extensive and evenly distributed.} Figure~\ref{fig:1} illustrates the detailed geographic distribution, showing that data points are present on all continents except Antarctica. The geographic distribution is uniform and reasonable, and can adequately represent the features of surface water body worldwide. Therefore, models trained on our dataset are expected to have stronger generalizability in the geographical dimension.
  \vspace{-0.5em}
  \item \textbf{The temporal span of image acquisition is broad.} The range of acquisition time of data spans from 2011 to 2022, and each year contains a certain amount of data. Models trained on this dataset exhibit greater temporal generalization ability.
  \vspace{-0.5em}
  \item \textbf{Inclusion of a diverse type of surface water landscapes}, as depicted in Figure~\ref{fig:3}. These include, but are not limited to, \textit{lakes and rivers in the forest, grassland, field, shrub, bare area, and urban area}, \textit{pools, glacial lakes, and water in the special scenario}. This wide types of water bodies serves as a representation of various geographic landscapes, land cover conditions, water body shapes, and color tone types, thus providing a comprehensive reflection of the diversity of the global surface water system.
\end{itemize}

In summary, the above five advantages drive \textit{GLH-water} dataset to be unique and advanced.

\subsection{Dataset splits}
To ensure that the training and test data are roughly equally distributed, we randomly select 80\% of the original images as the training set, 10\% as the validation set, and 10\% as the test set. We will make all original images publicly available with annotations. From the geographical distribution range shown in Figure~\ref{fig:1}, we can find that the validation and test sets are randomly distributed in various regions of the world, which can well reflect the actual performance of the model trained by this dataset.

\section{Methods on \textit{GLH-water} dataset}
\label{sec4}
\begin{figure}
	\centering
		\includegraphics[scale=.36]{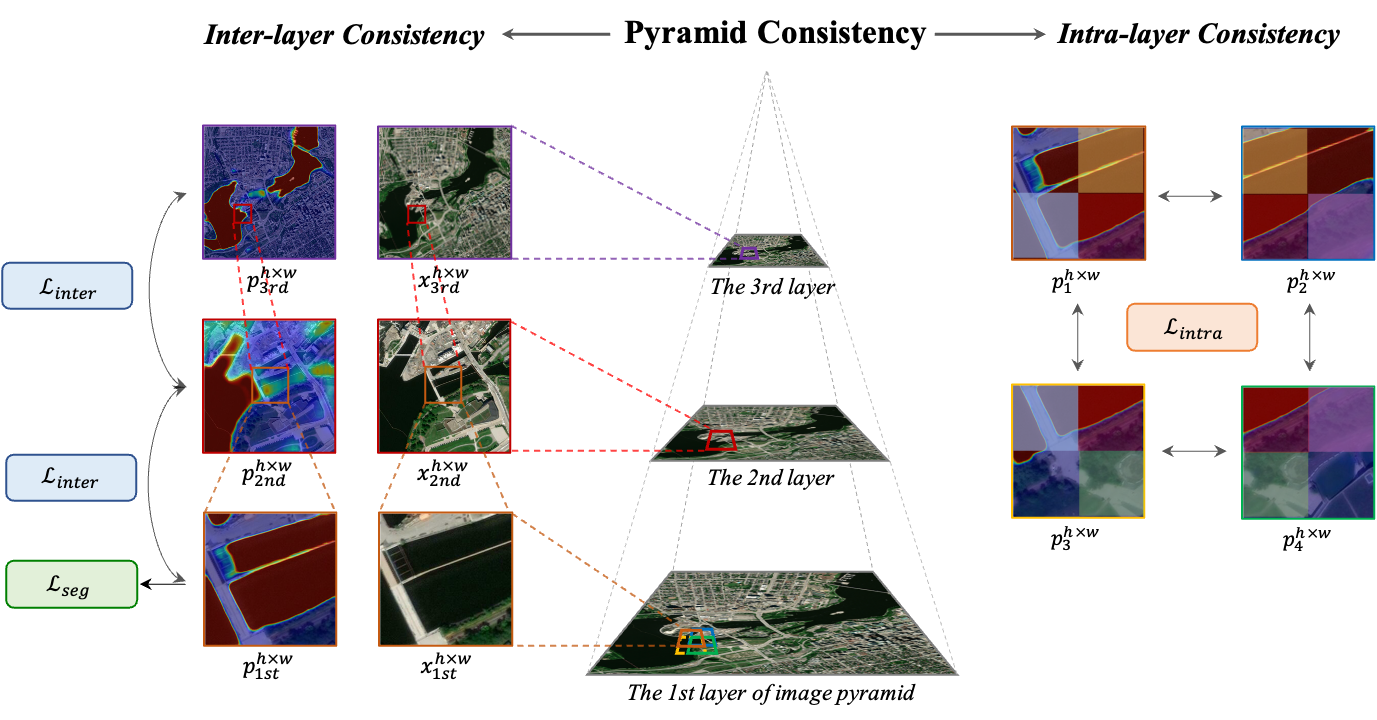}
	    \caption{\textbf{An overview of our proposed strong baseline with the PCL.}}
	\label{fig:4}
\vspace{-1.0em}
\end{figure}
\subsection{Baseline models}
\label{sec4.1}
Many models are developed to consider the characteristics of water bodies in VHR satellite images in the field of remote sensing, as outlined in Section~\ref{sec2.2}. We choose three representative models (\ie, MECNet \cite{zhang2021rich}, MSResNet \cite{dang2021msresnet}, and MSCENet \cite{kang2021multi}) as baseline models. In the realm of computer vision, numerous sophisticated semantic segmentation models are perpetually created, which can easily be adapted for satellite images. We use five advanced models, namely PSPNet \cite{zhao2017pyramid}, DeepLab v3+ \cite{chen2018encoder}, HRNet \cite{wang2020deep}, \etc., to construct the benchmark results. In addition, given the VHR and the large size of images in \textit{GLH-water}, we also evaluate the state-of-the-art semantic segmentation algorithms specifically designed for ultra-high resolution images (\ie, FCtL \cite{Li_2021_ICCV}, MagNet \cite{Huynh_2021_CVPR}, and ISDNet \cite{Guo_2022_CVPR}).

\subsection{A strong baseline with the PCL}
We develop a competitive strong baseline with the new PCL that is specifically designed to explore the detection performance of surface water bodies in large-size VHR satellite images. The pyramid consistency encompasses the intra-layer consistency (\ie, visual field consistency between pyramidal layers) and the inter-layer consistency (\ie, spatial consistency within a pyramidal layer), as illustrated in Figure~\ref{fig:4}. To construct the image pyramid, we downsample each original large-size image $X^{H \times W}$ at varying rates $\left\{\sigma_{i},i=n \right\}$, resulting in a multi-layer representation. Considering computational cost and efficiency, we adopt downsampling rates of 1, 1/5, and 1/25, generating an image pyramid comprising three layers.

\begin{figure}[t]
  \centering
   \includegraphics[width=\linewidth]{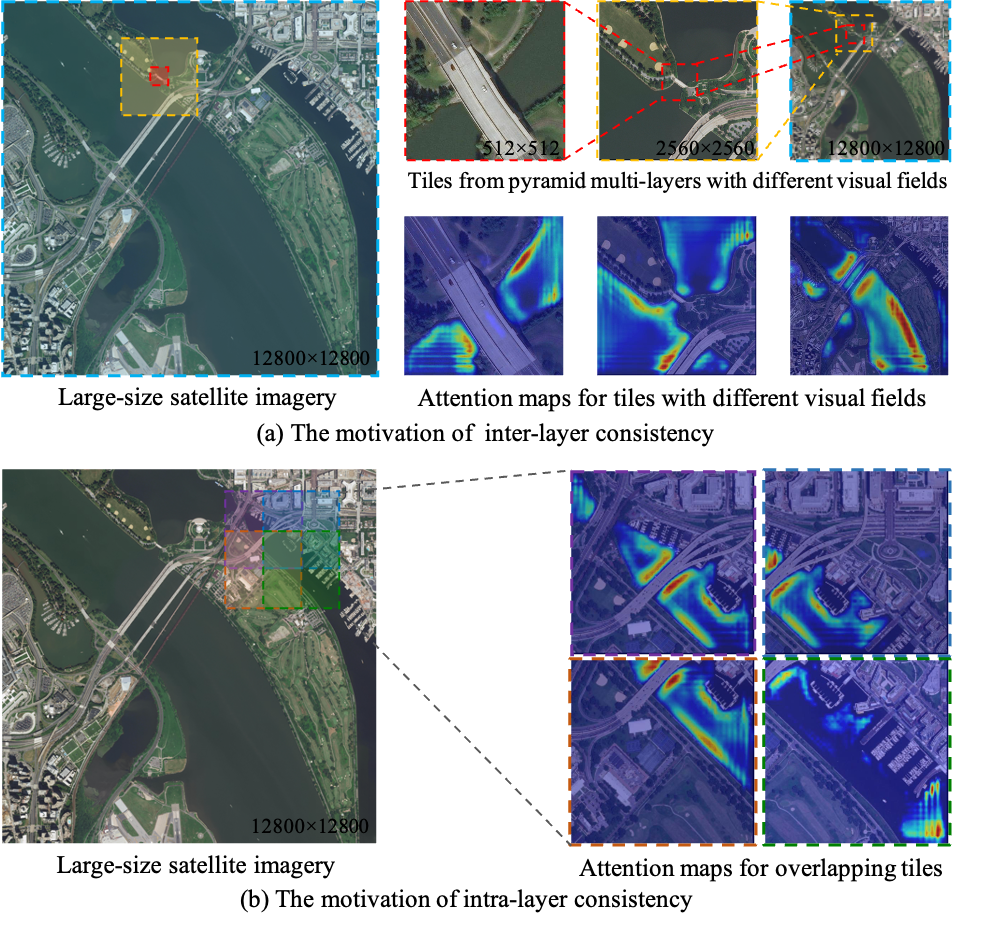}
   \caption{\textbf{The motivation of pyramid consistency.} Attention maps of tiles with different visual fields and overlapping tiles are distinct in the same areas. Attention maps is obtained by the GradCAM \cite{selvaraju2017grad}, using the ResNet-50 model trained on the \textit{GLH-water} dataset.}
   \label{fig:5}
\vspace{-1.5em}
\end{figure}

\textbf{Inter-layer consistency.} As mentioned in \cite{min2022peripheral}, the human brain may be influenced by the varying sizes of the visual field being observed, potentially resulting in divergent interpretations. The differences in attention maps of tiles with distinct visual fields displayed in Figure~\ref{fig:5}(a) show that the model is influenced by context information associated with the visual field. Motivated by this idea, we propose the inter-layer consistency loss to calculate the discriminative variances of the model resulting from dissimilarities in the visual range of patches. Specifically, we define the small tiles in the original image (\ie tiles located in the first layer of the pyramid) $x_{1st}^{h\times w}$ as the fundamental units and establish inter-layer tile groups $\left\{x_{1st}, x_{2nd}, x_{3rd}\right\}^{h \times w}$ by upwardly mapping the corresponding tiles from various layers. It is pertinent to note that while the size of each tile within the tile group remains consistent and same, the visual field they contained is gradually increasing as shown in Figure~\ref{fig:4}. The tile groups are trained by the encoder $\mathcal{E}$, decoder $\mathcal{D}$, and classifier $\mathcal{C}$ to obtain the corresponding $sigmoid$ normalized confidence maps $\left\{p_{1st}, p_{2nd}, p_{3rd}\right\}^{h \times w}$. Minimizing the differences between same regions in the confidence maps, which are caused by differences in the visual field, alleviates the visual field bias that arises due to limited contextual information.

\textbf{Intra-layer consistency.} Slicing the original image for processing may lead to the loss of contextual information and interdependence between adjacent tiles, which potentially disrupts the topological continuity of water bodies in remote sensing images with a large size. Figure~\ref{fig:5}(b) implies that the models exhibit differentiated attention for adjacent tiles with overlaps. We argue that enforcing consistency of the overlapping region on tiles with different contextual information helps to resume the continuity of water bodies. Based on this challenge and motivation, we further develop the intra-layer consistency loss to effectively model the continuous relationship between neighboring tiles and compensate for the information gap induced by image slicing. Specifically, we define four adjacent and overlapping tiles as an intra-layer tile group $\left\{x_1, x_2, x_3, x_4\right\}^{h \times w}$, and their spatial relationships are depicted in Figure~\ref{fig:4}. All tile pairs $ \left\{\left(x_i, x_j\right), 1 \leqslant i<j \leqslant 4 \right\} $ are processed by the encoder $\mathcal{E}$, decoder $\mathcal{D}$, and classifier $\mathcal{C}$ to obtain $sigmoid$ normalized confidence map pairs $\left(p_i, p_j\right)$. The overlapping part between them is employed to calculate the intra-layer consistency loss.

\textbf{Loss function.} Inspired by the focal loss \cite{lin2017focal}, we modify and present a novel consistency loss function to effectively calculate both inter-layer consistency $\mathcal{L}_{inter}$ and intra-layer consistency $\mathcal{L}_{intra}$ abovementioned. Overall optimization objective function can be defined as follows:
\begin{equation}
\setlength\abovedisplayskip{3pt}
\setlength\belowdisplayskip{3pt}
\label{deqn_ex1}
\mathcal{L}_{total}=\mathcal{L}_{seg}+\alpha_{inter} \mathcal{L}_{inter}+\alpha_{intra} \mathcal{L}_{intra},
\end{equation}
where $\mathcal{L}_{seg}$ denotes the regular semantic segmentation loss using binary cross entropy loss $\ell_{bce}$. $\alpha_{inter}, \alpha_{intra}$ are trade-off weights. $\mathcal{L}_{inter},\mathcal{L}_{intra}$ are formulated as
\begin{equation}
\setlength\abovedisplayskip{3pt}
\setlength\belowdisplayskip{3pt}
\label{deqn_ex2}
\begin{split}
\mathcal{L}_{inter}=\frac{1}{w \times h} \sum_{}^{w \times h} \left(1-p_{1st}\right)^r(1-\lambda) y_{1st} \ell_{2}\left({p}_{1st}, \tilde{p}_{2nd}\right) \\ +\lambda p_{1st}^r\left(1-y_{1st}\right) \ell_{2}\left({p}_{1st}, \tilde{p}_{2nd}\right) \\
+\frac{1}{w \times h} \sum_{}^{w \times h} \left(1-p_{1st}\right)^r(1-\lambda) y_{1st} \ell_{2}\left({p}_{1st}, \tilde{p}_{3rd}\right) \\ +\lambda p_{1st}^r\left(1-y_{1st}\right) \ell_{2}\left({p}_{1st}, \tilde{p}_{3rd}\right),
\end{split}
\end{equation}
\begin{equation}
\setlength\abovedisplayskip{3pt}
\setlength\belowdisplayskip{3pt}
\label{deqn_ex3}
\begin{split}
\mathcal{L}_{intra}=\frac{1}{w \times h} \sum_{}^{w \times h} \sum_{1 \leqslant i<j \leqslant 4} \left(1-\tilde{p}_i\right)^r(1-\lambda) \tilde{y}_i \ell_{2}\left(\tilde{p}_{i}, \tilde{p}_{j}\right) 
\\ +\lambda \tilde{p}_i^r\left(1-\tilde{y}_i\right) \ell_{2}\left(\tilde{p}_{i}, \tilde{p}_{j}\right),
\end{split}
\end{equation}
where $y_{1st}$ denotes the tiles and corresponding binary annotations in first layer. The value of 1 denotes water type in the given pixel, while the value of 0 indicates non-water type. ${p}_{1st}$ denotes the confidence map of tiles in first layer, and $\tilde{p}_{2nd},\tilde{p}_{3rd}$ represent the confidence maps of the overlapping regions in the inter-layer tile group of other layers (after upsampling). $ \ell_{2}\left({p}_{1st}, \tilde{p}_{2nd}\right)=\left\|{p}_{1st}-\tilde{p}_{2nd}\right\|_2^2 $ calculates the square of the euclidean distance of ${p}_{1st}, \tilde{p}_{2nd}$. Similarly, $\tilde{p}_{i},\tilde{p}_{j}$ represent the confidence map of the overlapping regions in the intra-layer tile group. $r, \lambda$ are tunable focusing parameters, which help the model to focus on learning hard-to-distinguish samples.

\section{Benchmark and experiment}
\subsection{Setup}
\noindent\textbf{Implementation details.} To ensure the fairness of the evaluation, the batchsize of baseline models, except ultra-high resolution segmentation methods, is set to 8 with a single NVIDIA TITAN RTX GPU. The initial learning rate adopts 10$^{-4}$. We crop the original image without overlap to 512 $\times$ 512 pixels and use the SGD optimizer with momentum 0.9, decayed weight 5×10$^{-4}$ to train baseline models for 15 epochs under a poly learning rate scheduler. To reduce training costs and achieve faster convergence, we uniformly use ImageNet \cite{deng2009imagenet} pre-trained backbones. For the implementation of ultra-high resolution segmentation methods, we crop the image to 2,560 $\times$ 2,560 pixels and set suitable parameter and hyperparameter settings refer to the settings in their paper to ensure that trained models converge. 

Due to GPU memory limitations, the batchsize of PCL is set to 4. The other hyperparameters are the same as those of baseline methods. The image pyramid is constructed in three layers, and the default sampling rates are set to 1, 1/5, and 1/25 according to experimental experience. The tile size of the first pyramid layer is 512 $\times$ 512 pixels to calculate the segmentation loss and the PCL. The trade-off weights $\alpha_{inter}$ and $\alpha_{intra}$ in Eq.~\eqref{deqn_ex1} are both set to 1.0. $r, \lambda$ are set to 2 and 0.2, respectively.

\noindent\textbf{Evaluation metrics.} We use the intersection-over-union (IoU) and F1-score metrics to evaluate the quantitative performance of detecing surface water, following previous related work. In addition, we also use Frames Per Second (FPS) and GPU Memory to evaluate the computational efficiency and consumption of different models.

\subsection{Evaluation results}
As described in Section~\ref{sec4}, we evaluate 12 of the popular methods shown in Table \ref{tab:table3}. The accuracy of generic semantic segmentation models is overall higher than that of models designed specifically for surface water body detection. In addition, the state-of-the-art methods performing on the \textit{Cityscapes} \cite{cordts2016cityscapes}, \textit{DeepGlobe} \cite{demir2018deepglobe}, and \textit{Inria} \cite{maggiori2017can} dataset (\ie, MagNet \cite{Huynh_2021_CVPR} and ISDNet \cite{Guo_2022_CVPR}) cannot achieve satisfactory performance on the \textit{GLH-water} dataset, which proves that surface water detection in large-size VHR satellite imagery is challenging and still needs further development. Furthermore, our proposed PCL outperforms other methods by leveraging the multi-layer field of visual information present in large-size images and the topological continuity of water bodies. Nonetheless, it suffers from low efficiency and high computational cost, which are common issues faced by other ultra-high resolution segmentation methods (\ie, FCtL \cite{Li_2021_ICCV} and ISDNet \cite{Guo_2022_CVPR}). Thus, striking a balance between accuracy and efficiency should be considered a crucial research priority in this task.

Table \ref{tab:table4} demonstrates the consistent improvement of our proposed strong baseline compared to the fair baseline approach across different segmentation model settings. This indicates that our approach is an effective pipeline and is driving progress in this task. Detailed visualization results will be presented in the supplementary material.

\begin{table}[]
\renewcommand\arraystretch{1.2}
\resizebox{84mm}{28mm}
{\begin{tabular}{cccccc}
\toprule
Method      & Backbone                                                      & IoU(\%) ($\uparrow$) & F1-score(\%) ($\uparrow$) & FPS ($\uparrow$) & Memory(MB) ($\downarrow$) \\ \midrule
\multicolumn{6}{l}{\textit{\textbf{Segmentation methods specifically proposed for surface water detection}}}               \\
MECNet \cite{zhang2021rich}      & -                                                             & 44.67         & 61.75         & 3.44    & 15749             \\
MSResNet \cite{dang2021msresnet}    & Res-34                                                        & 69.76         & 82.18         & 4.03     & 5429             \\
MSCENet \cite{kang2021multi}     & Res2-50                                                       & 74.81         & 85.58         & 2.60    & 6347            \\ \midrule
\multicolumn{6}{l}{\textit{\textbf{Generic segmentation methods in Computer Vision}}}                                      \\
FCN8s \cite{long2015fully}      & VGG-16                                                        & 73.66    & 84.83         & 6.70    & 8294             \\
PSPNet \cite{zhao2017pyramid}      & Res-50                                                        & 75.19    & 85.84         & 5.98    & 5451             \\
DeepLab v3+ \cite{chen2018encoder} & Res-50                                                        & 79.80         & 88.76         & 4.48    & 4973             \\
HRNet-48 \cite{wang2020deep}   & -                                                             & 78.60         & 88.01         & 3.03    & 8513            \\
STDC-1446 \cite{fan2021rethinking}   & -                                                             & 75.82         & 86.25               & \textbf{26.50}    & \textbf{2435}             \\ \midrule
\multicolumn{6}{l}{\textit{\textbf{Ultra-high Resolution Segmentation methods}}}                                           \\
MagNet \cite{Huynh_2021_CVPR}      & FPN-Res-50                                                    & 62.77         & -              & 13.33     &  4796           \\
FCtL \cite{Li_2021_ICCV}        & FCN8s-VGG16                                                   & 74.92         & 85.66              & 0.112    & 8192             \\
ISDNet \cite{Guo_2022_CVPR}     & DeepLab v3-Res-18                                             & 53.04    &  -             & 2.09    & 13608            \\ \midrule
\textbf{Our PCL}        & \begin{tabular}[c]{@{}c@{}}PSPNet\\ -Res-50\end{tabular} & \textbf{82.26}         & \textbf{90.27}              & 1.34    & 14503            \\ \bottomrule
\end{tabular}}
\vspace{0.0em}
\caption{\textbf{Benchmark results} of baseline models and our proposed model on \textit{GLH-water} \texttt{test} set. FPS and Memory are measured in training settings with batchsize=2.}
\label{tab:table3}
\vspace{-1.0em}
\end{table}

\begin{table}[h]
\renewcommand\arraystretch{1.2}
\resizebox{84mm}{20mm}
{\begin{tabular}{cccc}
\toprule
Method   & Seg model                           & IoU (\%)               & F1-score (\%)          \\ \midrule
Baseline & \multirow{3}{*}{FCN8s-VGG16}        & 73.66                       & 84.83                       \\
FCtL     &                                     & 74.92                  & 85.66                  \\
\textbf{Our PCL}     &                                    & \textbf{75.78} \footnotesize{(\textcolor{blue}{$+$2.12})}                       & \textbf{86.23} \footnotesize{(\textcolor{blue}{$+$1.4})}                       \\ \midrule
Baseline & \multirow{2}{*}{PSPNet-Res-50}      & 75.19                  & 85.84                  \\
\textbf{Our PCL}     &                                     & \textbf{82.26} \footnotesize{(\textcolor{blue}{$+$7.07})} & \textbf{90.27} \footnotesize{(\textcolor{blue}{$+$4.43})}  \\ \hline
Baseline & \multirow{2}{*}{DeepLab v3+-Res-50} & 79.80                  & 88.76                  \\
\textbf{Our PCL}     &                                     &   \textbf{81.33} \footnotesize{(\textcolor{blue}{$+$1.53})}                      &  \textbf{89.70} \footnotesize{(\textcolor{blue}{$+$0.94})}                       \\ \bottomrule
\end{tabular}}
\vspace{0.0em}
\caption{Performance of baseline with divergent backbone or the existing method with the fair backbone and our proposed model on \textit{GLH-water} \texttt{test} set. The results show that our model outperforms common models when using various segmentation models.}
\label{tab:table4}
\vspace{-1.0em}
\end{table}

\begin{figure*}[h]
	\centering
		\includegraphics[scale=.31]{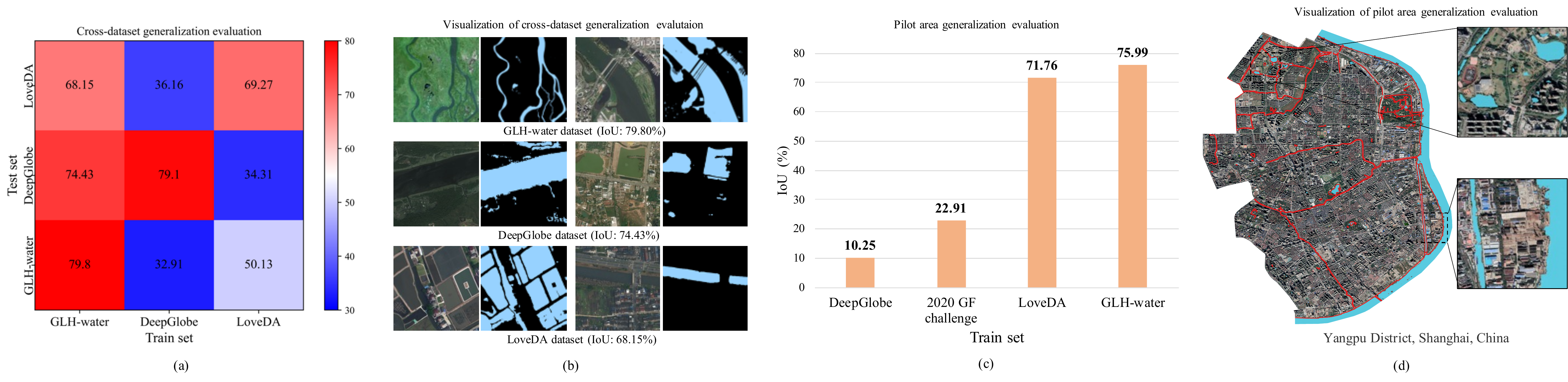}
            \vspace{-1.5em}
	    \caption{\textbf{Generalization experiment results on \textit{GLH-water} dataset.} (a) IoU (\%) of cross-dataset evaluation. (b) Visualization results of cross-dataset evaluation. The displayed results in three datasets are all predicted by the model trained on \textit{GLH-water}. (c) IoU (\%) of pilot area evalution with the HRNet-48 models trained on different datasets. (d) Visualization results of the model trained by our \textit{GLH-water} on the pilot area. \textcolor{red}{Red lines} represent ground truth, and \textcolor{cyan}{cyan} masks are predictions.}
	\label{fig:6}
\vspace{-1.0em}
\end{figure*}

\subsection{Ablation study on the strong baseline}
\noindent \textbf{Effectiveness of components in PCL.} Exps. II and III in Table~\ref{tab:table5} show that both key components of PCL (\ie, $\mathcal{L}_{inter}$ and $\mathcal{L}_{intra}$) outperform the baseline by a large margin (+5.89\% and +5.50\%), and their combination can further improve the performance of the model (Exp. VII).

\noindent \textbf{Effectiveness of loss function of PCL.} We conduct an ablation study using the vanilla L2 loss function to measure the pyramid consistency (Exp. IV), and find that the loss function we designed (Eqs.~\eqref{deqn_ex2} and \eqref{deqn_ex3}) has superior capacity to facilitate learning on hard-to-distinguish samples, thereby resulting in performance improvement.

\noindent \textbf{Impact of the number of image pyramid layers.} We observe that if we only apply two layers of the image pyramid to participate in training, our PCL will bring a performance improvement of over 4.9\% (Exps. V and VI). When building three layers of the image pyramid, we can see that there is a 7.07\% improvement (Exp. VII), indicating that more layers being considered may be more beneficial.

\begin{table}[]
\setlength\fboxsep{0pt}
\centering
\renewcommand\arraystretch{1.2}
\resizebox{82mm}{18mm}
{\begin{tabular}{c|c|c|c}
\toprule
ID & Configuration & IoU (\%) & \multicolumn{1}{c}{$\Delta$ (\%)} \\ \midrule
I & Baseline ($\mathcal{L}_{seg}$)            & 75.19        & -      \\
II & $\mathcal{L}_{seg} + \mathcal{L}_{inter}$             & 81.08        & \textcolor{blue}{$+$5.89}      \\
III & $\mathcal{L}_{seg} + \mathcal{L}_{intra}$             & 80.69        & \textcolor{blue}{$+$5.50}       \\
IV & Vanilla $\ell_{2}$             & 71.46        & \textcolor{red}{$-$3.73}                               \\
V & Without the 2rd layers of the image pyramid             & 80.13        & \textcolor{blue}{$+$4.94}      \\
VI & Without the 3nd layers of the image pyramid             & 81.37        & \textcolor{blue}{$+$6.18}       \\\midrule
VII & \textbf{Our PCL}             & \textbf{82.26}        & \textcolor{blue}{$+$7.07}     \\ \bottomrule
\end{tabular}}
\vspace{0.5em}
\caption{\textbf{Ablation study on key components of strong baseline} (seg model: PSPNet-Res-50). \texttt{Vanilla $\ell_{2}$} means using vanilla L2 loss to measure the pyramid loss rather than loss function (Eqs.\eqref{deqn_ex2} and \eqref{deqn_ex3}) we designed.}
\label{tab:table5}
\vspace{-1.0em}
\end{table}

\section{Generalization experiments on \textit{GLH-water}}
\subsection{Cross-dataset generalization evaluation}
Considering the similar resolution, data source, and large image size, we choose the \textit{LoveDA} \cite{wang2021loveda} and \textit{DeepGlobe} dataset \cite{demir2018deepglobe} to implement the cross-dataset generalization evaluation. Following the data split of \cite{wang2021loveda,Guo_2022_CVPR}, we use DeepLab v3+-Res-50 as the segmentation model to train the models and evaluate the cross-dataset performance.

As shown in Figure~\ref{fig:6}(a), there is little difference between the results of the model trained on \textit{GLH-water} and the model trained on \textit{LoveDA} on the \textit{LoveDA} test set (68.15\% vs. 69.27\%). However, the performance of the model trained on \textit{LoveDA} is significantly diminished when transferred directly to the \textit{GLH-water} test set (50.13\% vs. 79.80\%). A similar situation occurrs between \textit{DeepGlobe} and \textit{GLH-water}. Results in Figure~\ref{fig:6}(a) and (b) confirm the strong generalization of our \textit{GLH-water}.

\subsection{Pilot area generalization evaluation}
Providing data support for future VHR global surface water mapping is one of the motivations for constructing the \textit{GLH-water}. We select the Yangpu District of Shanghai, China, which is independent of the dataset and annotated by experts, as a pilot area (60.61 $km^{2}$) to further discuss the generalization of \textit{GLH-water}.

Based on the results presented in Figure~\ref{fig:6}(c) and (d), it is evident that the model trained on \textit{GLH-water} exhibits superior performance (75.99\%) in the surface water mapping task in the pilot area, surpassing the models trained on other datasets. These findings suggest that \textit{GLH-water} holds significant potential for global-scale VHR surface water mapping, owing to its strong generalization.

\section{Qualitative results}
\label{sec:seca2}
\begin{figure*}[htbp]
\centering
		\includegraphics[scale=.82]{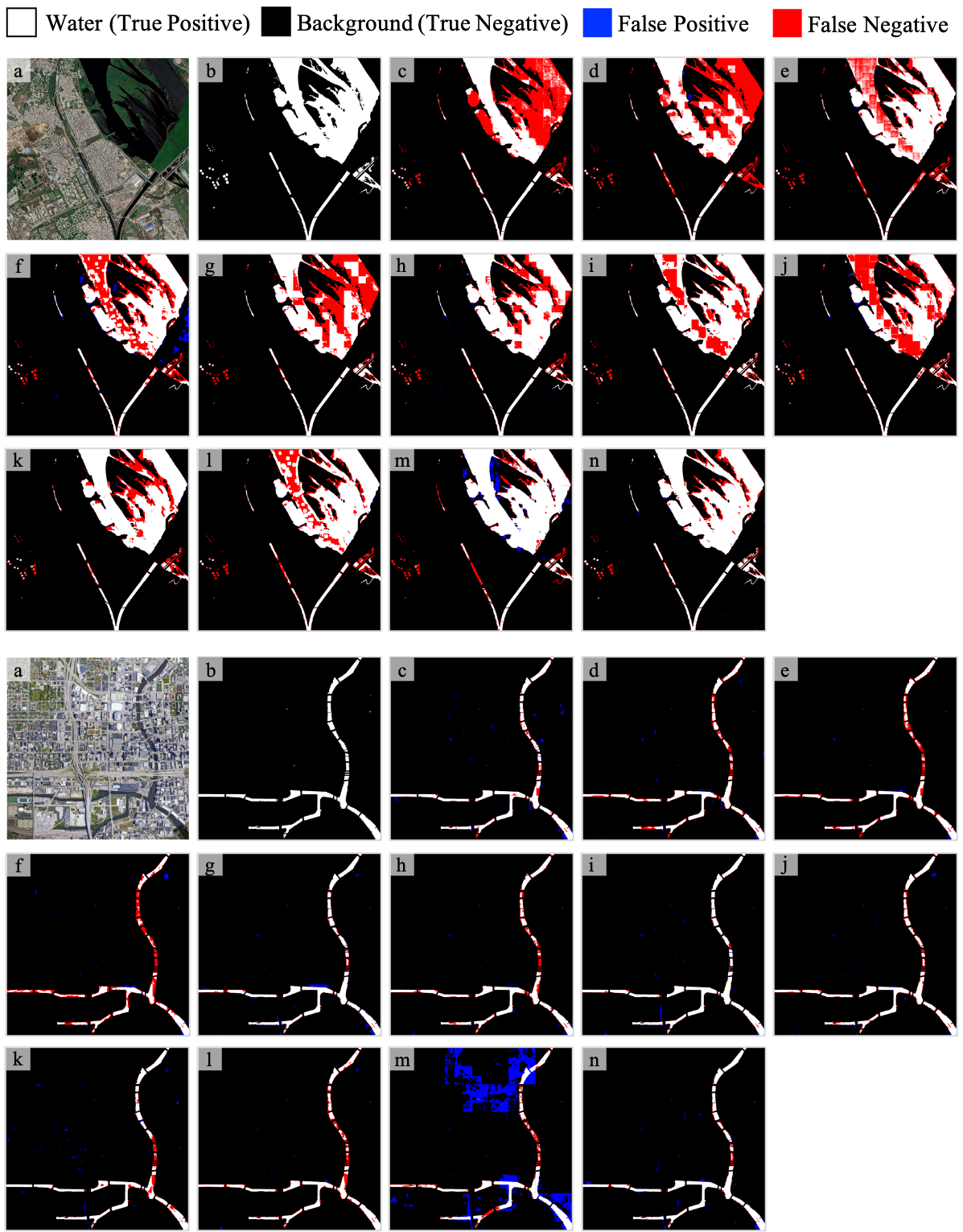}
	    \caption{\textbf{Qualitative results on the \textit{GLH-water} \texttt{test} set.} (a) Original image. (b) Ground truth. (c)-(e) Results of MECNet, MSResNet, and MSCENet, respectively. (f)-(j) Results of FCN8s, PSPNet, DeepLab v3+, HRNet-48, and STDC-1446, respectively. (k)-(m) Results of MagNet, FCtL, and ISDNet, respectively. our PCL can preserve more complete lakes and rivers, with much fewer commission errors (\textcolor{blue}{false positive}) and omission errors (\textcolor{red}{false negative}), as shown in (n).}
	\label{fig:sup1}
\end{figure*}

\begin{figure*}[htbp]
\centering
		\includegraphics[scale=.82]{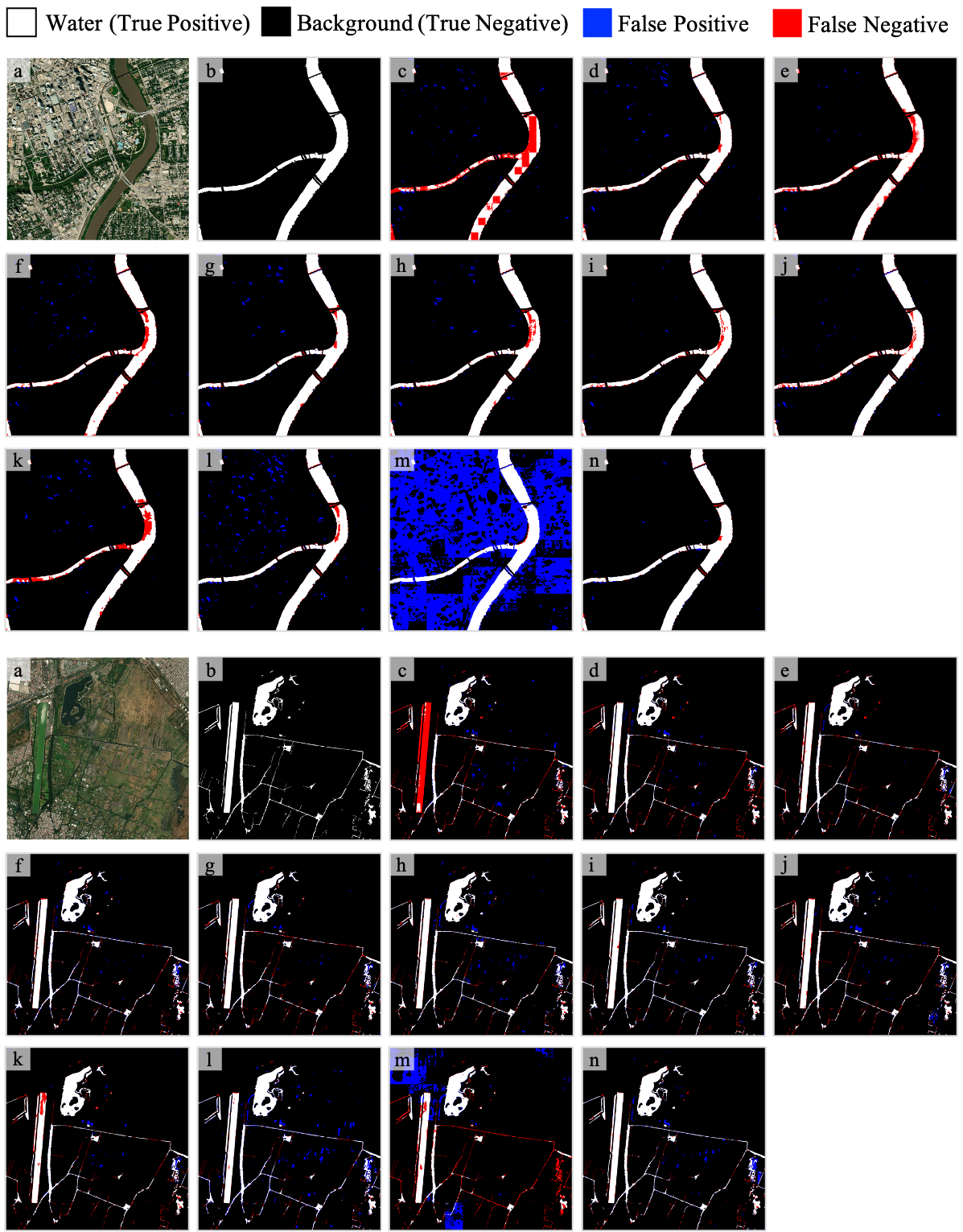}
	    \caption{\textbf{Qualitative results on the \textit{GLH-water} \texttt{val} set.} (a) Original image. (b) Ground truth. (c)-(e) Results of MECNet, MSResNet, and MSCENet, respectively. (f)-(j) Results of FCN8s, PSPNet, DeepLab v3+, HRNet-48, and STDC-1446, respectively. (k)-(m) Results of MagNet, FCtL, and ISDNet, respectively. our PCL can preserve more small and slim rivers, with much fewer commission errors (\textcolor{blue}{false positive}) and omission errors (\textcolor{red}{false negative}), as shown in (n).}
	\label{fig:sup2}
\end{figure*}

\noindent \textbf{Qualitative results on \textit{GLH-water} \texttt{test} and \texttt{val} sets.} Figures~\ref{fig:sup1} and \ref{fig:sup2} show the surface water detection visualization results of the 12 methods included in the benchmark on the \texttt{test} set and \texttt{validation} set of \textit{GLH-water}, respectively. Compared with other results, our PCL can better preserve more continuous and complete surface water bodies and reduce commission errors, as shown in (n) of Figure~\ref{fig:sup1}. In addition, our PCL can detect small and slim rivers, as shown in (n) of Figure~\ref{fig:sup2}.

\begin{figure*}[htbp]
\centering
		\includegraphics[scale=.80]{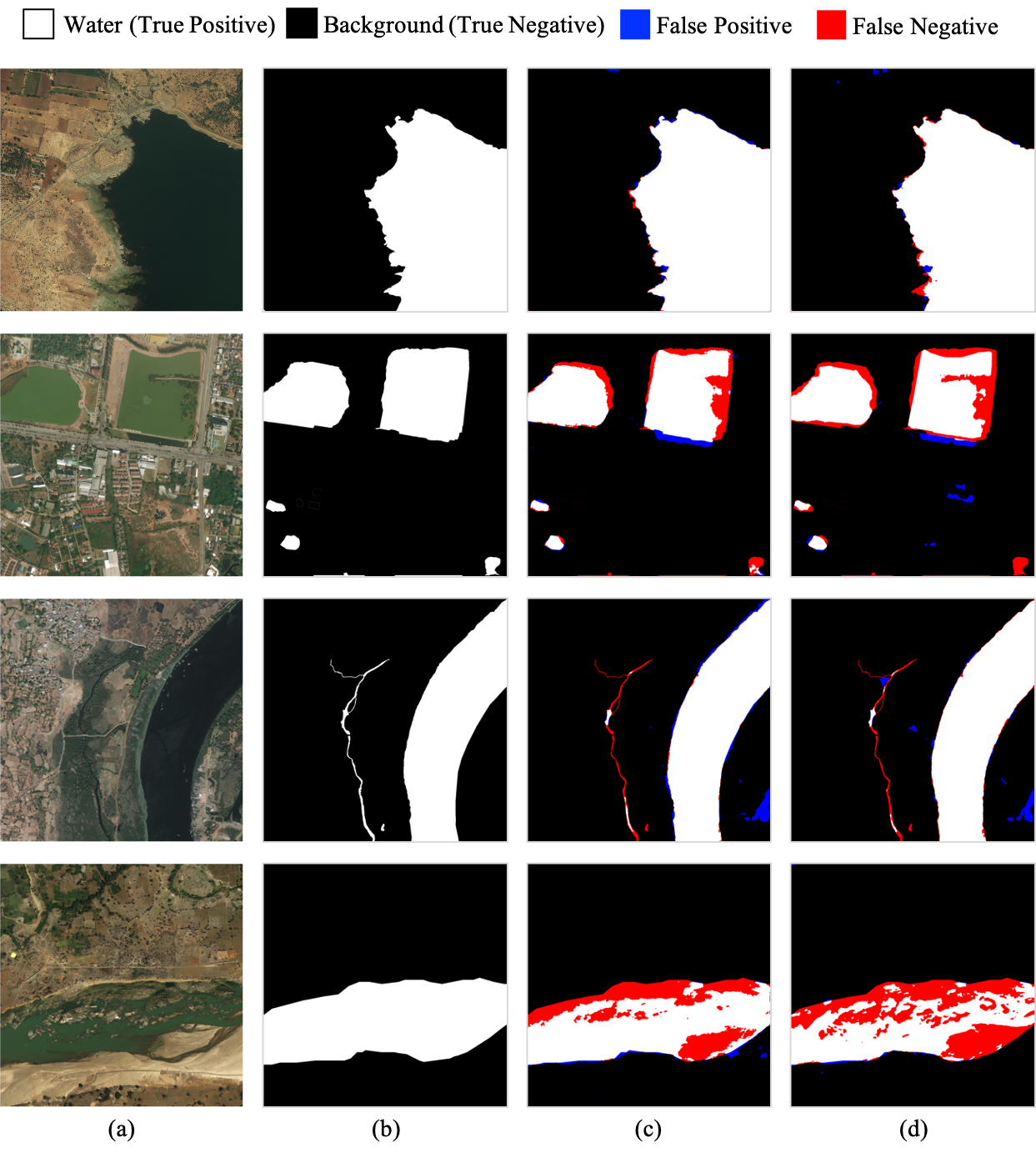}
	    \caption{\textbf{Qualitative results on the \textit{DeepGlobe} \texttt{test} set.} (a) Original image. (b) Ground truth. (c) Results of model trained on \textit{DeepGlobe} train set (IoU: 79.1\%). (d) Results of model trained on our \textit{GLH-water} train set (IoU: 74.43\%). As in the comparison of quantitative results, the performance of (d) and (c) is not significantly different, indicating that \textit{GLH-water} has strong generalization.}
	\label{fig:sup3}
\end{figure*}

\begin{figure*}[htbp]
\centering
		\includegraphics[scale=.80]{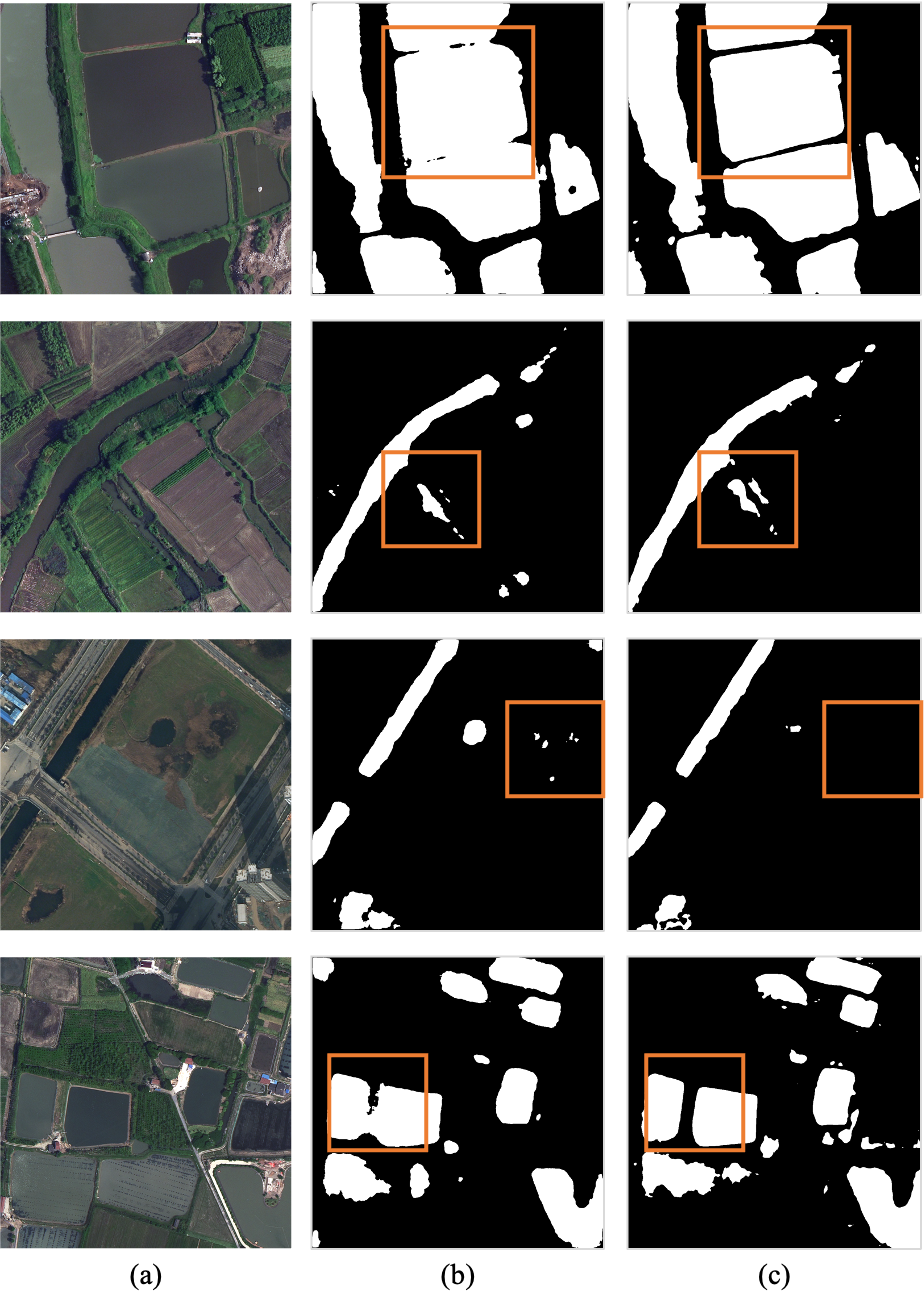}
	    \caption{\textbf{Qualitative results on the \textit{LoveDA} \texttt{test} set.} (a) Original image. (b) Results of model trained on \textit{LoveDA} train set (IoU: 69.27\%). (c) Results of model trained on our \textit{GLH-water} train set (IoU: 68.15\%). (Ground truth is not available and quantitative results are obtained from the online leaderboard: \url{https://codalab.lisn.upsaclay.fr/competitions/421}.) The detection details of surface water bodies shown in (c) are better than those in (b), as shown in \textcolor{orange}{orange boxes}.}
	\label{fig:sup4}
\end{figure*}

\noindent \textbf{Qualitative results about cross-dataset generalization evaluation.} Figures~\ref{fig:sup3} and \ref{fig:sup4} display the visualization results on the \textit{DeepGlobe} \cite{Demir_2018_CVPR_Workshops} and \textit{LoveDA} \cite{wang2021loveda} \texttt{test} sets of models trained on different datasets, respectively. As shown in the quantitative results presented in our paper (IoU: 74.43\% vs. 79.1\% and 68.15\% vs. 69.27\%), thanks to the strong generalization of \textit{GLH-water}, the model trained on the \textit{GLH-water} \texttt{train} set achieves performance on the \textit{DeepGlobe} and \textit{LoveDA} \texttt{test} sets that is not significantly different from the performance of the model trained on their respective \texttt{train} sets.

\begin{figure*}[htbp]
\centering
		\includegraphics[scale=.95]{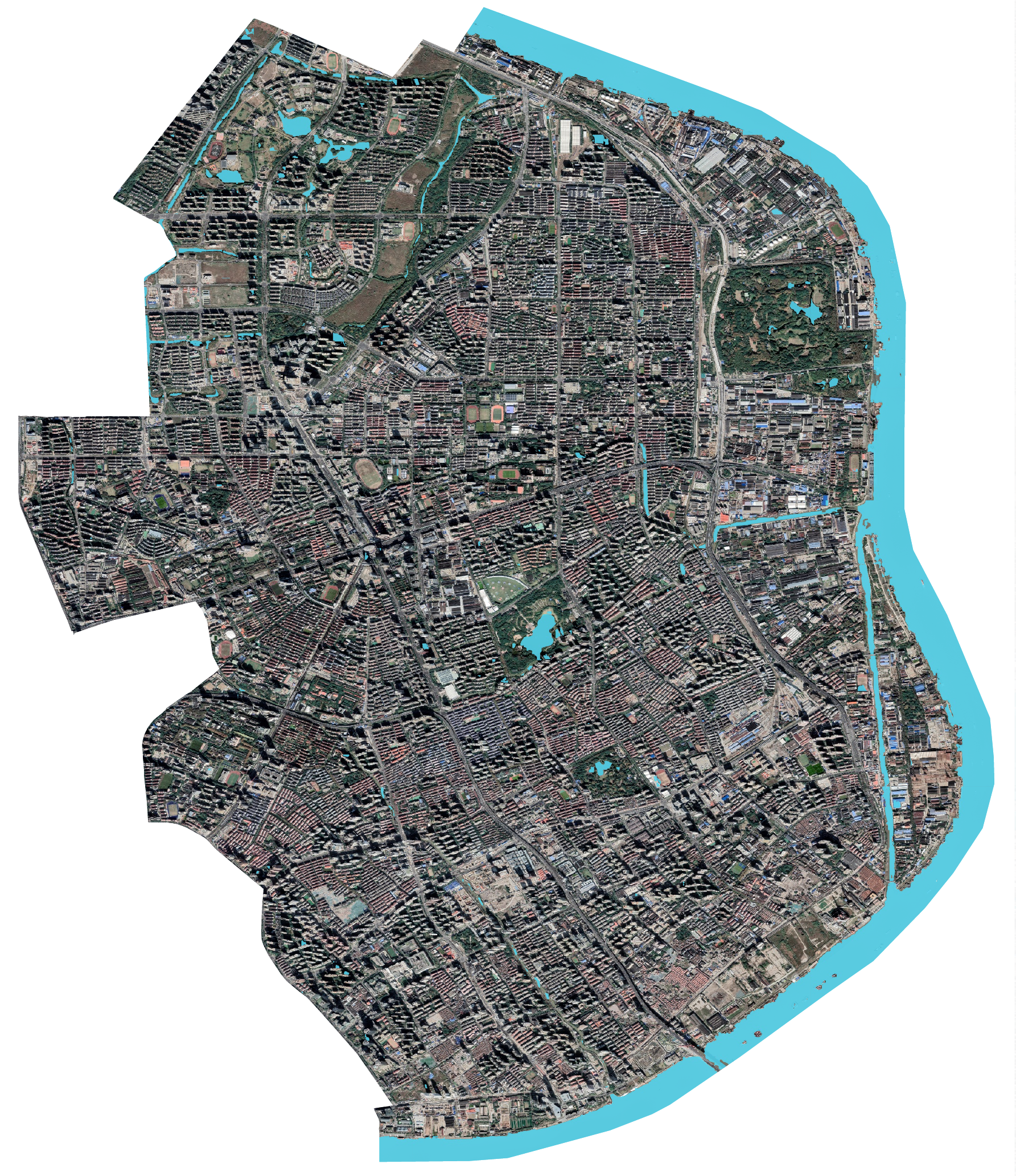}
	    \caption{\textbf{Qualitative results on the pilot area (Yangpu District, Shanghai, China).} \textcolor{cyan}{Cyan} masks are predicted by the model trained on \textit{GLH-water} \texttt{train} set.}
	\label{fig:sup5}
\end{figure*}

\noindent \textbf{Qualitative results on pilot area.} Figure~\ref{fig:sup5} shows the performance of our model trained on \textit{GLH-water} dataset for surface water detection in an out-of-sample pilot area (Yangpu District, Shanghai, China). The model trained on \textit{GLH-water} exhibits superior performance (75.99\%) in the surface water mapping task in this pilot area, surpassing the models trained on \textit{DeepGlobe} \cite{Demir_2018_CVPR_Workshops}, \textit{2020 GF challenge} \cite{sun2021automated}, and \textit{LoveDA} \cite{wang2021loveda} datasets. Additionally, Figure~\ref{fig:sup5} also demonstrates that our model's ability to detect slender streams in urban scenes and resist the interference of building shadow is limited. These limitations should be considered as future research priorities for surface water detection from large-size VHR satellite imagery.

\section{Conclusion}
We present a global large-scale dataset for surface water detection in large-size VHR satellite imagery which is first publicly available dataset in this task. Unlike exisiting datasets, we collect 250 large-size satellite images containing various surface water scenes across the whole earth and carefully annotate their masks. Considering the advantages of \textit{GLH-water} over other datasets, we believe that this dataset is more appropriate for practical applications and more challenging. Additionally, we build a benchmark to evaluate advanced segmentation models in the fields of remote sensing and computer vision. we aslo propose a strong basline with PCL that is a promising research pipeline to push this task forward.

In the future, we hope the \textit{GLH-water} with strong generalization will not only evaluate the progress of algorithms, but also provide data to support major studies such as global surface water mapping and even global water resource conservation and management.
\\ \hspace*{\fill} \\
\\ \hspace*{\fill} \\
\\ \hspace*{\fill} \\
\\ \hspace*{\fill} \\
\\ \hspace*{\fill} \\
\\ \hspace*{\fill} \\
\\ \hspace*{\fill} \\
\\ \hspace*{\fill} \\
\\ \hspace*{\fill} \\
\\ \hspace*{\fill} \\
\\ \hspace*{\fill} \\
\\ \hspace*{\fill} \\
\\ \hspace*{\fill} \\
\\ \hspace*{\fill} \\
\\ \hspace*{\fill} \\
\\ \hspace*{\fill} \\
\\ \hspace*{\fill} \\
\\ \hspace*{\fill} \\
\\ \hspace*{\fill} \\
\\ \hspace*{\fill} \\
\\ \hspace*{\fill} \\
\\ \hspace*{\fill} \\
\\ \hspace*{\fill} \\
\\ \hspace*{\fill} \\
\\ \hspace*{\fill} \\
\\ \hspace*{\fill} \\
\\ \hspace*{\fill} \\
\\ \hspace*{\fill} \\
\\ \hspace*{\fill} \\
\\ \hspace*{\fill} \\
\\ \hspace*{\fill} \\
\\ \hspace*{\fill} \\
\\ \hspace*{\fill} \\
\\ \hspace*{\fill} \\
\\ \hspace*{\fill} \\
\\ \hspace*{\fill} \\
\\ \hspace*{\fill} \\
\\ \hspace*{\fill} \\
\\ \hspace*{\fill} \\
\\ \hspace*{\fill} \\
\\ \hspace*{\fill} \\
\\ \hspace*{\fill} \\
\\ \hspace*{\fill} \\
\\ \hspace*{\fill} \\
\\ \hspace*{\fill} \\
\\ \hspace*{\fill} \\
\\ \hspace*{\fill} \\

{\small
\bibliographystyle{ieee_fullname}
\bibliography{main}
}

\end{document}